\documentclass[letterpaper, 10 pt, conference]{ieeeconf}  

\IEEEoverridecommandlockouts                              

\usepackage{cite}

\usepackage{graphicx}
\usepackage{amsmath}
\usepackage{subfigure} 
\usepackage{hyperref}
\usepackage[dvipsnames]{xcolor}
\usepackage{algorithm} 
\usepackage{algpseudocode} 
\usepackage{siunitx}

\usepackage[deletedmarkup=sout, authormarkup=superscript]{changes}
\definechangesauthor[name={Kelly}, color=RedViolet]{KM}
\definechangesauthor[name={Joris}, color=orange]{JD}
\definechangesauthor[name={Gaoyuan}, color=Green]{GL}

\title{\LARGE \bf
A Task-Efficient Reinforcement Learning Task-Motion Planner \\for Safe Human-Robot Cooperation 
}

\author{Gaoyuan Liu$^{1,2}$, Joris de Winter$^{1,3}$,  Kelly Merckaert$^{1,3}$, Denis Steckelmacher$^{4}$, \\ Ann Nowe$^{4}$, and Bram Vanderborght$^{1,2}$
\thanks{This work was supported by the \textit{Flemish} Government under the program \textit{Onderzoeksprogramma Artifici\H ele Intelligentie (AI) Vlaanderen} and China Scholarship Council (CSC) and euROBIN (No. 101070596).}
\thanks{$^{1}$ Authors are with the Department of Mechanical Engineering, Vrije Universiteit Brussel, Brussels, Belgium. {\tt\small gaoyuan.liu@vub.be}}%
\thanks{$^{2}$ Authors are affiliated to imec, Belgium}%
\thanks{$^{3}$ Authors are affiliated to Flanders Make, Belgium}%
\thanks{$^{4}$ Ann Nowe and Denis Steckelmacher are with the Artificial Intelligence (AI) Lab, Vrije Universiteit Brussel, Brussels, Belgium.}%
}

\begin{document}

\maketitle

\thispagestyle{empty}
\pagestyle{empty}

\begin{abstract}

In a Human-Robot Cooperation (HRC) environment, safety and efficiency are the two core properties to evaluate robot performance. However, safety mechanisms usually hinder task efficiency since human intervention will cause backup motions and goal failures of the robot. Frequent motion replanning will increase the computational load and the chance of failure. 
In this paper, we present a hybrid Reinforcement Learning (RL) planning framework which is comprised of an interactive motion planner and a RL task planner. The RL task planner attempts to choose statistically safe and efficient task sequences based on the feedback from the motion planner, while the motion planner keeps the task execution process collision-free by detecting human arm motions and deploying new paths when the previous path is not valid anymore. Intuitively, the RL agent will learn to avoid dangerous tasks, while the motion planner ensures that the chosen tasks are safe. The proposed framework is validated on the cobot in both simulation and the real world, we compare the planner with hard-coded task motion planning methods. The results show that our planning framework can 1) react to uncertain human motions at both joint and task levels; 2) reduce the times of repeating failed goal commands; 3) reduce the total number of replanning requests.


\end{abstract}

\section{INTRODUCTION}

Human-Robot Cooperation (HRC), in which human workers and collaborative robots have to share the workspace, is expected to boost dexterity and productivity of manufacturing processes, but this is often jeopardized by an unsafe or inefficient coordination between humans and robots.

To protect human workers, abundant motion planning algorithms enable the robot to steer the end-effector to the goal position while avoiding the surrounding obstacles in the joint space. For an environment which contains uncertainty such as human motions, an intuitive solution is to frequently replan paths to cope with the current obstacles' distribution \cite{park2012itomp, petti2005safe, park2013real}. However, task efficiency is usually neglected or sacrificed to improve safety, e.g. robot performance can be conservative when replanning is activated constantly, which will stall the robot to make progress on the task level. Moreover, such replanning algorithms usually have high computational load \cite{park2013real}. Therefore, it's difficult to find the sweet spot between safety and efficiency\cite{zhao2021efficient}.

\begin{figure}
    \centering
    \includegraphics[width=0.35\textwidth]{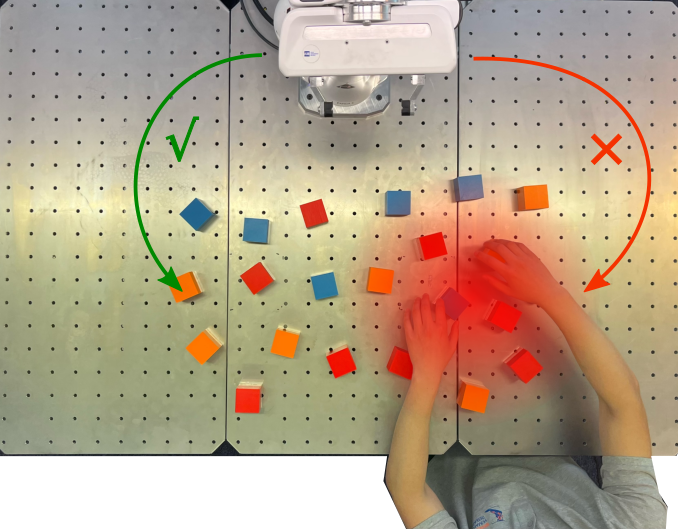}
    \vspace{-0.1cm}
    \caption{Human-Robot Cooperation (HRC) environment where the human and the robot share a workspace. The robot and the human share the same manipulation goal which is to pick up and pack all the objects on the table. On top of the motion planning in the joint space to avoid the human arms, the robot also needs to avoid the occupation in the task space where the human has higher possibility working in, which is depicted as the red zone. }
    \vspace{-0.3cm}
    \label{fig:intro}
\end{figure}

Recent research on Task and Motion Planning (TAMP) considers both symbolic task planning and geometric motion planning \cite{garrett2021integrated} by building hierarchical planners. Despite the achievement on quality task planning based on prior discrete and symbolic modeling of the domain, a robot can still face uncertainties that are difficult to predict. Consider a HRC scenario shown in Fig. \ref{fig:intro}, where the human and the robot work on homogeneous tasks in a shared workspace, each block represents a sub-task and the human and the robot share the job which is to pick and pack all the blocks to the bin. Beside interactively avoiding collisions with human arms, the robot can reduce replanning requests and repeating failures by avoiding congested areas. However, the human's occupation in the workspace is variable. Therefore, it is hard to model the change of the congested areas and plan the detour paths which avoid the congested areas. In this sense, Reinforcement Learning (RL) shows its ability to improve the robots' statistic performance\cite{mnih2015human}.


In this work, we consider safety and efficiency in a HRC environment as a TAMP problem and solve them comprehensively. The solution given in this paper is twofold. First, we implement an interactive motion planner, named re-RRT* on the robot manipulator to avoid human arms. The number of replanning requests is firstly reduced by checking the current path and only replanned when necessary. Second, a RL task planner is integrated to choose task positions that are both currently accessible and statistically obstacle-sparse in the long run. The RL task planner intends to further reduce the number of replanning requests and the task failure count for finishing all the tasks. 

The two modules are interconnected. The motion planner provides a safe environment for the RL exploration process, the safe RL environment allows the robot to learn and to finish the job at the same time. Also, the number of replanning requests will be fed into the RL agent as reward data. On the other hand, the RL task planner tries to give a rational goal reference according to the prediction of the obstacle statistics. The contributions of this work include:

The remainder of the paper is organized as follows. Section \ref{sec:related_work} presents an overview of the related work. Section \ref{sec:methodology} details the methods and concepts in our framework which can fit in the generic RL environment formulation and our main contribution of a novel safety layer. In Section \ref{sec:experiment_and_results}, we analyze the learning performance with our framework and demonstrate the obtained results. Section \ref{sec:conclusion} concludes the paper.



\section{RELATED WORK} \label{sec:related_work}


Abundant motion planning algorithms are developed in the previous work. They can be categorized into two classes: sampling-based \cite{sucan2012open} and optimization-based \cite{kalakrishnan2011stomp}. To avoid dynamic obstacles, an intuitive solution is requesting new trajectories frequently based on the current situation \cite{park2012itomp, petti2005safe, park2013real, bruce2002real}. Optimization-based algorithms need several iterations to find an optimal or sub-optimal trajectory. Therefore, frequently replanning on-the-fly will cause severe delay. Park et al. \cite{park2013real} improve the performance of the optimization-based algorithms by utilizing better hardware. Sampling-based algorithms such as Rapidly-exploring Random Tree (RRT)\cite{karaman2011anytime} can respond to requests fairly fast, however, may result in trajectories that are prone to undesirable jerky motions \cite{ota2019trajectory}. Moreover, replanning is generally computationally expensive and with significant delays. To handle such dilemmas, real-time replanning interleaves planning with execution so that the robot may decide to compute only partial or sub-optimal plans in order to avoid delays in the movement \cite{park2012itomp,hauser2012responsiveness}. 
In \cite{pupa2021safety}, the aforementioned dilemma is solved by only requesting new trajectories under certain circumstances. 

The recently flourishing machine learning research gives new perspectives to handle the HRC safety issue. Markov Decision Process (MDP) approximation algorithms are used to improve HRC efficiency by predicting human intentions \cite{bandyopadhyay2013intention}. Neural-network guided motion planners are introduced in \cite{qureshi2020motion, zucker2008adaptive}. The RL algorithms are inherently suitable for robot tasks. A model-free RL agent can learn feasible trajectories under dynamic constraints. Dynamic obstacle avoidance is achieved with Deep Reinforcement Learning (DRL) in \cite{el2020towards}. Besides DRL works as the motion planner, a recurrent neural network (RNN) model is combined as a motion predictor to improve the legibility of the motion planner \cite{zhao2020actor}. However, training a RL planner from scratch requires a lot of time before converging to good solutions. The hybrid methods usually results in better performance.
In \cite{ota2019trajectory}, the sample-based algorithm guides the RL agent to learn dynamically-feasible trajectories.
A two-fold structure is developed in \cite{zhao2020actor} which contains a DRL local motion planner and a goal selector in order to improve task efficiency. In their works, even the trajectories for the different goal positions are labeled and trained simultaneously. The trained RL planner can only plan for a single static goal position. A high-level task planner is added on top of a low-level trajectory planner in \cite{park2019planner}, using a Q-Learning algorithm to calculate robot reactions according to the prediction of human motions. We use a similar hybrid strategy to tackle the task-efficiency problem.

Task planning and scheduling is studied with variant objectives. Given the criteria such as productivity \cite{tsarouchi2017human}, human fatigue \cite{zhang2022cycle}, ergonomics\cite{el2019task}, capability \cite{raatz2020task} and efficiency \cite{ferreira2021scheduling}, task planning and scheduling strategies attempt to allocate tasks between the human and the robot. In our work, we emphasize the connection between the task planning and safety. 
Integrated 
TAMP is formalized as a research area where robot motions are planned on different levels, considering both logical and geometric constraints \cite{garrett2021integrated}. Task planning introduces a new angle to improve HRC safety and efficiency by adapting robot behaviours and minimizing mutual interference\cite{mansouri2021combining}. A hierarchical task and motion planning strategy is proposed to improve HRC safety \cite{faroni2020layered}, but the task layer doesn't take into account motion properties. A RL combined method is introduced in \cite{le2021hierarchical}, the inverse RL is used as a human motion predictor while the TAMP algorithm Logic-Geometric Programming (LGP) is combined to plan minimal interference tasks.


\section{METHODOLOGY}\label{sec:methodology}
\begin{figure}[t]
    \centering
    \vspace{2 mm}
    \includegraphics[width=0.4\textwidth]{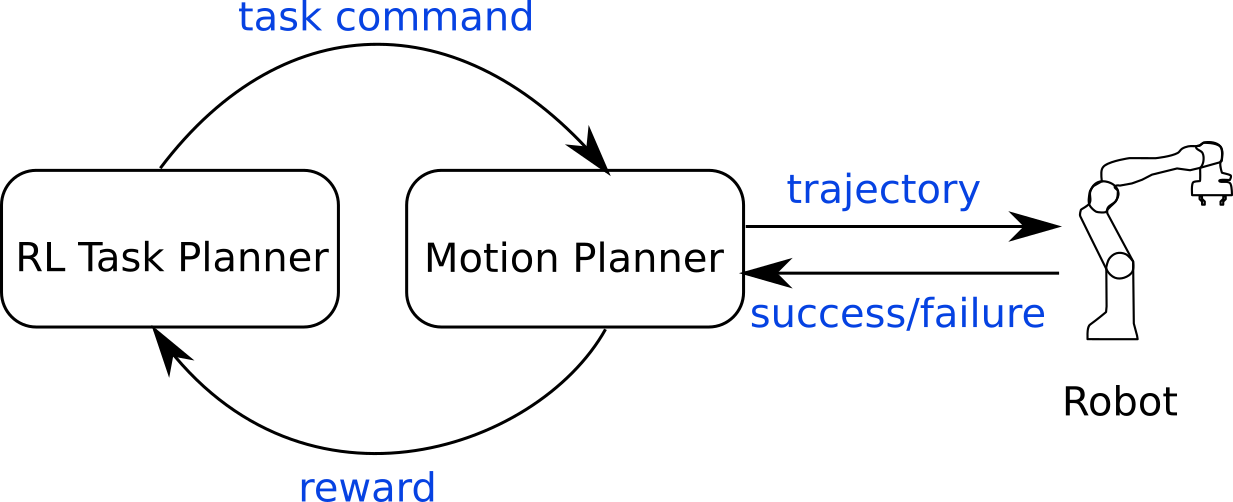}
    \caption{The planning framework is comprised of a task planner and a motion planner. The RL task planner gives task commands to the motion planner, the motion planner attempts to refine task commands by replanning safe trajectories. Combining the number of replanning requests and weather the task is accessible, the motion planner evaluates the task planner's decisions by giving rewards, which also helps the RL agent to improve the ability to make better decisions.}
    \vspace{-0.5cm}
    \label{fig:method_structure}
\end{figure}

The framework proposed in this paper is a hybrid RL task-motion planner, shown in Fig. \ref{fig:method_structure}, containing a low level motion planner and a high level RL task planner. The two planners function interactively. The motion planner tries to go to the goal position requested by the RL task planner, while avoiding uncertain human motions by constantly checking the current trajectory and replanning new trajectories when necessary. In the meantime, the RL task planner attempts to choose end-effector positions that are less likely to be occupied by the human arms, and are therefore statistically safer and more efficient. In other words, the motion planner provides a safe learning environment for the RL task planner, while the RL task planner reduces the probability that the motion planner needs to be activated.

\subsection{Interactive re-RRT* Motion Planning}
\begin{figure}
    \centering
    \vspace{2 mm}
    \subfigure[]{\includegraphics[width=0.23\textwidth]{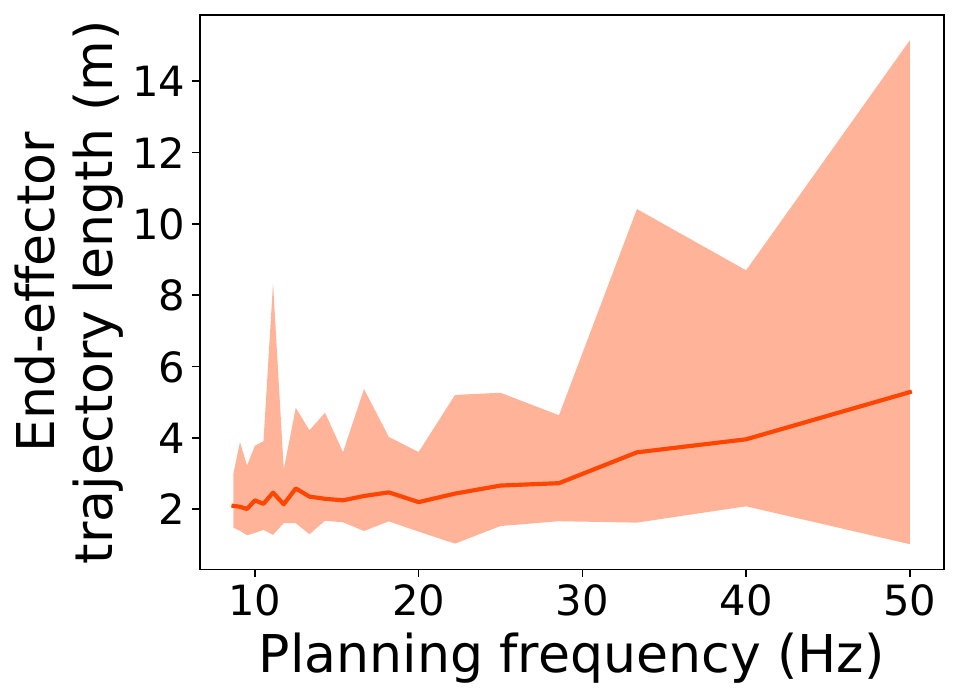}\label{fig:replanner_dilemma_a}} 
    \subfigure[]{\includegraphics[width=0.22\textwidth]{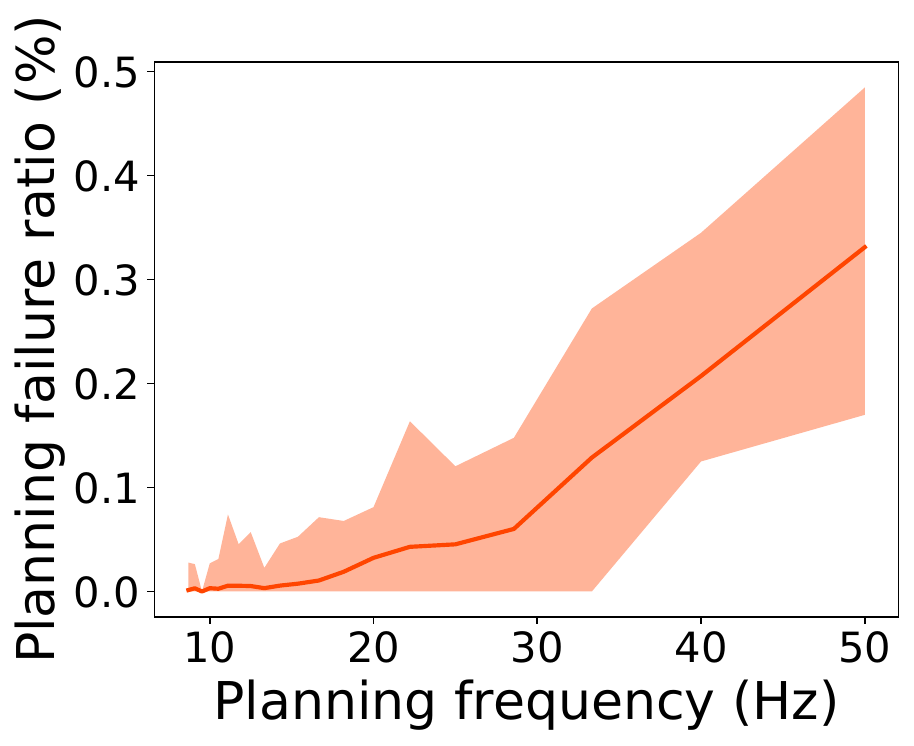}\label{fig:replanner_dilemma_b}} 
    \caption{The tradeoff between the planning frequency and optimality. (a) The end-effector trajectory elongates when the replanning frequency increases, because following frequently-changing sub-optimal trajectories will make the end-effector constantly change moving directions. (b) The planning failure ratio also increases when the replanning frequency increases, because of the lack of time for the planner to find the solutions. On the other hand, reducing the replanning frequency will cause the severe motion delay, which will deteriorate the safety.}
    \label{fig:replanner_dilemma}
\end{figure}

Robots must frequently adjust their motions on-the-fly to respond to uncertain human motions. Replanning algorithms solve the dilemma between efficiency and optimality by adapting the time devoted to planning \cite{hauser2012responsiveness}. However, the replanning dilemma between the replanning rate and the planner performance in a dynamic environment is still a challenge. To illustrate the dilemma, we implement a replanning strategy in a dynamic scenario shown in Fig. \ref{fig:task_setting_sim}. It uses RRT* as the basic planner, which gives new trajectories frequently. The robot has to go to the goal position while avoiding the moving human arms. The variation of the end-effector trajectory length and the planning failure ratio with the replanning frequency are illustrated in the Fig. \ref{fig:replanner_dilemma}. As shown in the figure, the end-effector trajectory length and the planning failure ratio deteriorate when the replanning frequency increases. The experiment results show that it is crucial to reduce the replanning frequency.
\begin{algorithm}
    \vspace{2 mm}
	\caption{RRT* Replanning  (\textbf{re-RRT*})} \label{alg: re-rrt*}
	\begin{algorithmic}[1]
    \State Set the end-effector goal position $\mathbf{a}=(x^{ee}, y^{ee}, z^{ee})$
    \State Initialize a path with RRT $\mathbf{\xi}_0 = [\mathbf{q}_0, \mathbf{q}_1, ..., \mathbf{q}_i,..., \mathbf{q}_M]$
    \State Predictive checking number $m$
    \State Observe the current end-effector position $\mathbf{p}$
    \State $\texttt{sucess} = \texttt{True}$
    \While {$\mathbf{p} \neq \mathbf{a}$ } %
        \State $\texttt{goal\_collision} = \mathbf{Collision\_Check}(\mathbf{q}_M)$
        \If{$\texttt{goal\_collision} = \texttt{True}$}
            \State Break
        \Else
            \For {$i < m$}
                \State $\texttt{waypoint\_collision} =  \mathbf{Collision\_Check}(\mathbf{q}_i)$
            \EndFor
            \If {$\texttt{waypoint\_collision} = \texttt{True}$}
                \State Replan a new path with RRT: $\mathbf{\xi}_{i+1} = [\mathbf{q}_0, \mathbf{q}_1, ..., \mathbf{q}_{i+1},..., \mathbf{q}_M]$
                
            \Else
                \State $\mathbf{\xi}_{i+1} = \mathbf{\xi}_{i}$
            \EndIf
            \State Follow the first waypoint in current path $\mathbf{\xi}_i[0]$
            \State Delete $\mathbf{\xi}_i[0]$  
        \EndIf
        \State Observe the current end-effector position $\mathbf{p}$
    \EndWhile
    \If{$\texttt{goal\_collision} = \texttt{True}$}
        \State Return $\texttt{sucess} = \texttt{False}$
    \Else
        \State Return $\texttt{sucess} = \texttt{True}$
    \EndIf
	\end{algorithmic} 
\end{algorithm}

\begin{figure}[h]
    \centering
    \vspace{2 mm}
    \includegraphics[width=0.3\textwidth]{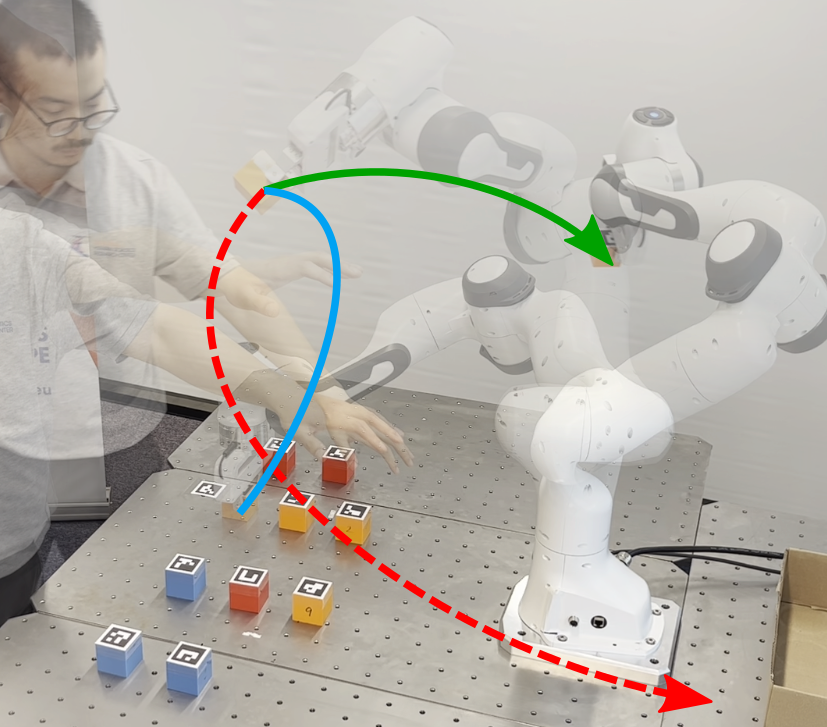}
    \caption{The effect of the re-RRT* motion planner. The blue line and red line represent the original trajectory, however, the human arms change position unexpectedly, the previous trajectory is not valid anymore after the blue section, a new successive trajectory (green) is planned to cope with the change.}
    \label{fig:motion_real_exp}
\end{figure}

In our work, we implement a RRT* replanning strategy (re-RRT*)\cite{connell2017dynamic}. As shown in Fig. \ref{fig:motion_real_exp}, instead of requesting new trajectories at a fixed frequency, re-RRT* constantly checks the current robot trajectory and only requests a new trajectory when the trajectory is occupied by an obstacle. We use the Flexible Collision Library (FCL) \cite{pan2012fcl} for The $\mathbf{Collision\_Check}$ function in the motion planner. Given the geometrical shape and position of an obstacle, and the joint angles of the robot, the collision checking function can respond in $2.5\cdot10^{-5}\si{.s}$. Therefore the essence of re-RRT* is that we use dense checking to avoid dense replanning. 

The re-RRT* algorithm can mitigate the forementioned dilemma: (1) the replanning frequency will be significantly reduced since unnecessary replanning requests will be avoided and (2) the collision checking in advance allows more time for planning and optimization.

Moreover, the replanning request time will be an important evaluation indicator (i.e. reward) for the policy of the RL task planner, in order to further reduce the replanning request time. The re-RRT* algorithm is presented in Algorithm \ref{alg: re-rrt*}.

\subsection{Reinforcement Learning Task Planning}
In this section, we formalize the task planning problem into the RL framework. In each learning step, the agent gives a command $\mathbf{a}_t$ in the Cartesian space that drives the robot end-effector to approach a new goal position. If the goal position is accessible, the nearest task around the goal location will be assigned to the robot. Next, the robot tries to execute (pick and drop) the assigned task with the re-RRT* motion planner, and gets the observation $\mathbf{o}$ and the reward $\mathbf{r}$ from the environment. To benefit from the decision-making ability of RL, the task planning problem can be formulated as a MDP, which is described as a tuple $(\mathcal{S}, \mathcal{A}, \mathcal{R}, \mathcal{T}, \mathcal{O}, \gamma)$, where $\mathcal{S}$ is the state space, $\mathcal{A}$ is the action space, $\mathcal{R}$ is the reward function, $\mathcal{T}$ is an unknown transition function, $\mathcal{O}$ is the observation space and $\gamma$ is a discount factor. Algorithm \ref{alg: rl-task-planner} is the pseudocode of the RL task planner. Below we describe the details of the observation space, the action space, and the reward function.

\subsubsection{Action Space}
We have set that the goal positions chosen by the RL task planner are always in a plane, therefore the action space contains the 2D position of the end-effector $[x_{ee}, y_{ee}]$. We use continuous location rather than a discrete task space as the action space because: (1) the size of the task space will shrink during the task finishing process; (2) using task space will cause dimensional curse when the number of tasks increases; (3) using a location space as action space implies that the model can be generalized to situations where the total number of tasks is different. After an end-effector position is chosen by the RL agent, the nearest task from that position will be assigned to the robot. This conversion allows us to convert the position decision to the task decision and the action space will not change with the number of tasks.

\subsubsection{Observation Space}
\begin{figure}
    \centering
    \vspace{2 mm}
    \subfigure[]{\includegraphics[width=0.3\textwidth]{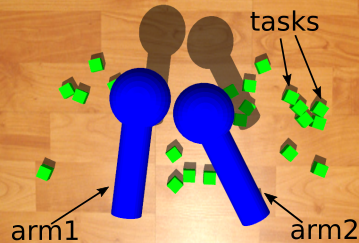}}
    \subfigure[]{\includegraphics[width=0.4\textwidth]{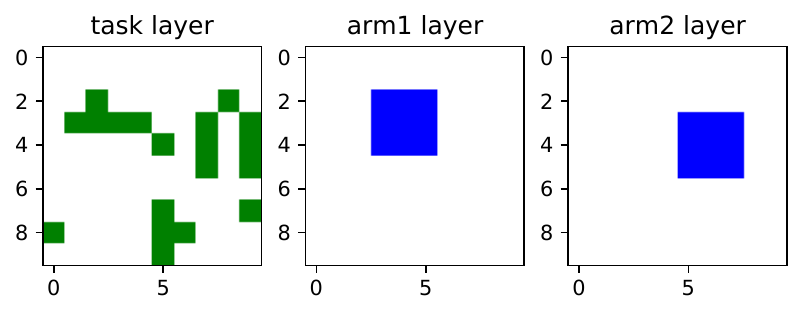}}
    \caption{The observation feature matrix visualization. (a) The task objects and human arms in the simulation, green boxes are the task objects and the blue cylinders-spheres objects indicate the human arms. (b) The visualization of the integer-encoding observed feature matrix, which is a $10\times10$, three-layer matrix. The green color indicates 1 in the matrix elements, the blue color indicates -1 in the matrix elements, while the white color indicates 0. }
    \label{fig:obs}
\end{figure}

The agent observes the current position of the human arms and the robot's remaining tasks. However, the number of remaining tasks changes during each episode. Also, in the cluttered tasks scenarios, the increase of the initial number of tasks will cause expanse of the observation space, i.e. curse of dimensionality.  To fix the size of the observation space and avoid the curse of dimensionality, we use a three layer feature matrix $F$ as observation input, where the three layers correspond to tasks and two human arms (arm1, arm2) respectively. One hot encoding method is used to mark the current 2D position of human arms $\mathbf{p}^\text{arm}_l = [x^\text{arm}_l, y^\text{arm}_l]$ and $\mathbf{p}^\text{arm}_r = [x^\text{arm}_r, y^\text{arm}_r]$, and the remaining tasks 2D position $\mathbf{p}_{\tau}^i = [x_{\tau}^i, y_{\tau}^i]$, where $i = [0...M]$ when there are $M$ remaining tasks. The feature matrix represents the 2D positions of the human arms and the tasks by dividing the work space into $10 \times 10$ sub-spaces. An example of the observation space is shown in Fig. \ref{fig:obs}. To be specific, $F(1, i, j) = 1$ when there are remaining tasks in the sub-space, $F(2, i, j) = -1$ when the left arm (arm 1) is in the sub-space and $F(3, i, j) = -1$ when the left arm (arm 2) is in the sub-space, the matrix elements remain 0 when there is no tasks or arms in the subspace.

It is worth noting that the integer-encoding observation is highly abstracted compared to the real geometry occupation truth. Such abstracted representation is sufficient and widely used for task planning in the TAMP field. That is, most TAMP methods use symbolic presentations in the task planner, while the motion planner takes care of the more accurate geometric properties \cite{jiang2019task}. 

\subsubsection{Reward Function}
\begin{algorithm}
	\caption{Reinforcement Learning Task Planner} \label{alg: rl-task-planner}
	\begin{algorithmic}[1]
		\For {$episode=1,2,\ldots$}
		    \State Reset the rest task list $\mathbf{T}$
			\For {$step=1,2,\ldots,N$}
			    \State Get action $\mathbf{a}_k = (x^{ee}_k, y^{ee}_k, z^{ee}_k)$ from the current policy $\mathbf{\pi}$
			    \State $d = ||(\mathbf{p}_{\tau_k}-\mathbf{p}_{ee})||^2$
			    \State $\texttt{goal\_collision} = \mathbf{Collision\_Check}(\mathbf{a}_k)$
				\If {$\texttt{goal\_collision} = \texttt{False}$}
				\State Assign the nearest task in the rest task list $\mathbf{\tau}_k \in \mathbf{T}$ 
				
				\State Execution $\texttt{success} = \textbf{re-RRT*}(\mathbf{\tau}_k)$
				\If {$\texttt{success} = \texttt{True}$}
				    \State $\texttt{task\_achieved} = \texttt{True}$
				    \State $\mathbf{T} \setminus \{ \mathbf{\tau}_k \}$ 
				    \Else
				    \State $\texttt{task\_achieved} = \texttt{False}$
				    \EndIf
				\Else 
				\State $\texttt{task\_achieved} = \texttt{False}$
				\EndIf
				\State Count the number of replanning requests $\mathbf{c}_k$ throughout the execution
				\State Reward $\mathbf{r}_k (\mathbf{c}_k, \texttt{task\_achieved}, d)$
				\State Get observation $\mathbf{o}_k$
    			\If {$\mathbf{T} = \emptyset$}
    			\State Break
    			\EndIf
			\EndFor
			\State Update the policy $\mathbf{\pi}$ with $\mathbf{r}_k$, $\mathbf{o}_k$
		\EndFor

	\end{algorithmic} 
\end{algorithm}
There are three evaluation indicators we use to optimize the RL task planner policy: (1) validation of the goal position; (2) replanning request time to go to the goal position; and (3) the distance between the goal position and the task. To optimize these three indicators, the reward function is given as follows:
\begin{equation} \label{eq1}
\begin{split}
r & =  \alpha_{0} \cdot [\texttt{task\_achieved}=\texttt{True}]\\
& -\alpha_{1} \cdot [\texttt{collision}=\texttt{True}] \\
& - \alpha_{2} \cdot \mathbf{c}_k - \alpha_{3} \cdot [d(\mathbf{p}_{ee}, \mathbf{p}_{\tau_k})]
\end{split}
\end{equation}
where $\alpha_{i}$ is the scaling weight, $\mathbf{p}_{ee}$ is the end-effector position, $\mathbf{p}_{\tau_k}$ is the task position which is nearest to the end-effector, and $\mathbf{c}_k$ is the total replanning time in one learning step. By maximizing the accumulated reward (i.e. return), the agent will be encouraged to plan goal positions that (1) are collision free; (2) need less replanning to achieve the task; and (3) are near the available task.

\subsection{Training}
We use a benchmarking policy gradient RL method, more specifically a Proximal Policy Optimization (PPO)\cite{schulman2017proximal}, to train the task planner.
Compared to its predecessor Trust Region Policy Optimization (TRPO) \cite{schulman2015trust}, PPO offers two key improvements to policy gradient methods: (1) surrogate objective includes a simple first order trust region approximation and (2) multiple epochs can be performed on collected data. These two modifications make PPO the most suitable strategy for our case: (1) the inconsistency of the sampling-based motion planners and the occasional simulation errors can cause abnormal rewards, which deviate the policy updates, and (2) the space of human motion is limited in the HRC scenario, thus multiple epochs can improve the sample efficiency.  


\section{Experiment and Results} \label{sec:experiment_and_results}
\subsection{Task Setting} \label{subsec:task_setting}
In our experiment, we assume a scenario where the human and the robot work together to pick up blocks and drop them into a bin. In this setting, human and robot have to share the workspace and cooperate simultaneously. To simplify the scenario, we make the assumption that the human arms appear in the workspace following a 2D (fixed height) Gaussian distribution in one episode. The mean of the Gaussian distribution is unpredictable and follows a uniform distribution. That is, the human arms move following the location sampled from: 
\begin{align*}
P(x^{\text{arm}}) = N(\mu_x, \sigma^2_x)\,, 
P(y^{\text{arm}}) = N(\mu_y, \sigma^2_y)\,.
\end{align*}
To make sure the human arms appearance will cover the whole 2D workspace, every episode will sample a new mean value of the Gaussian distribution from: 
\begin{align*}
P_{\mu_x} = U(x_\text{min}, x_\text{max})\,,
P_{\mu_y} = U(y_\text{min}, y_\text{max})\,,
\end{align*}
where $[x_\text{min}, x_\text{max}]$ and $[y_\text{min}, y_\text{max}]$ are the ranges of the workspace. 

\begin{figure}
    \centering
    \vspace{2 mm}
    \includegraphics[width=0.35\textwidth]{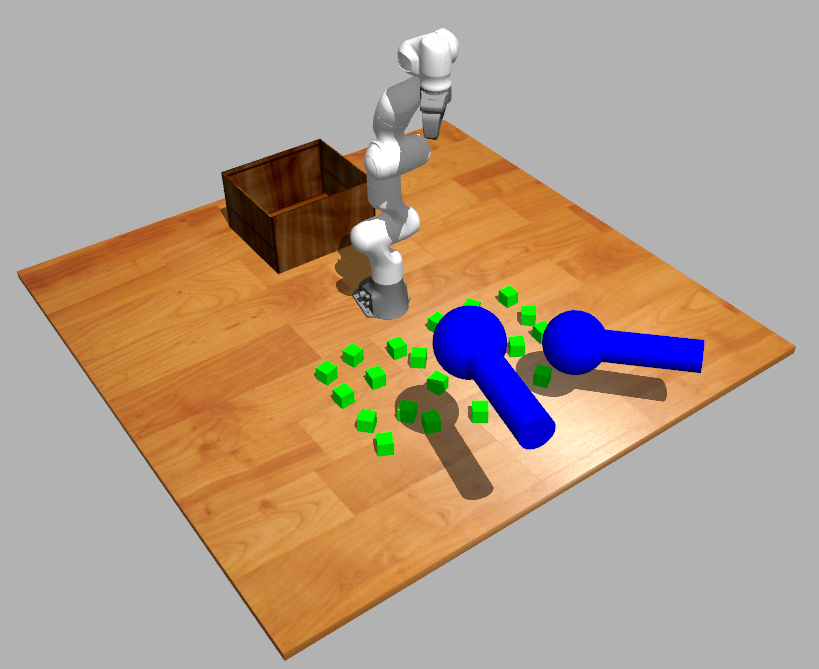}\label{fig:gazebo_setting}
    \caption{The simulation environment of the task setting. The green blocks are the tasks which need to be picked and dropped in the wooden bin behind the robot. The blue cylinder and sphere present the occupancy of the human hand.}
    \label{fig:task_setting_sim}
\end{figure}

The blocks are scattered in the workspace. The robot's job is to pick up and drop half of the blocks. The expected robot behaviour is that the robot can minimize the failure times and the number of replanning request time by avoiding the human working space.
We make the assumption that the human-arms occupation is the only reason that a task is inaccessible. Clustered and piled blocks are out of the scope of this paper. The scenario implementation in simulation is shown in Fig. \ref{fig:task_setting_sim}.

\subsection{Experiment Setting}
We deploy and test the proposed framework in simulation and real-world respectively. We use an Franka Emika Panda robot, in short Panda robot, a 7 DoF manipulator which is widely used for research purpose.
 
Our work connects the state-of-the-art robotic and RL software libraries. We use Robot Operate System (ROS) as the platform and Gazebo as the simulation environment to collect training data. MoveIt provides the basic RRT motion planner and integrates the FCL as the collision checker. OpenAI gym Stable-Baselines3 is the benchmarking RL library which provides advanced RL algorithms. To leverage the aforementioned software and libraries, we create the interface between the ROS and the OpenAI gym, that is, we build the RL training environment on ROS with the standard OpenAI gym environment formation. 

\begin{figure}
    \centering
    \vspace{2 mm}
    \subfigure[]{\includegraphics[width=0.35\textwidth]{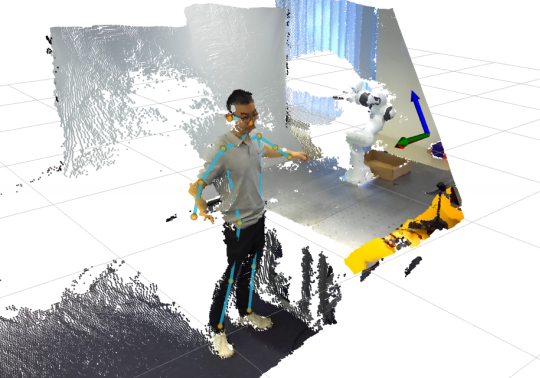}\label{fig:gazebo_setting}} 
    \subfigure[]{\includegraphics[width=0.35\textwidth]{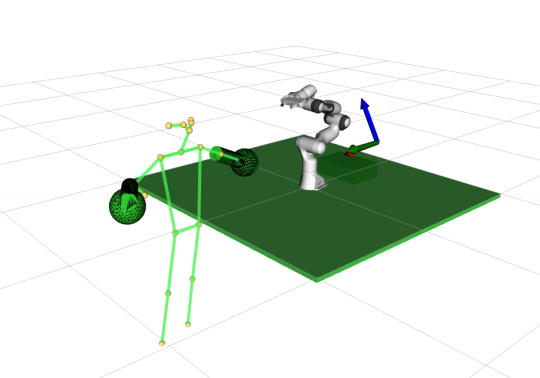}\label{fig:real_setting}} 
    
    \caption{The abstraction from point cloud data to obstacle information. (a) The point cloud captured by ZED2 camera, and the skeleton tracking data generated from the point cloud data. (b) The obstacle information we generated from skeleton tracking data, the green cylinder-sphere shape and the green table present human arm obstacle and table obstacle for the motion planner.}
    \label{fig:zed}
\end{figure}

As for the physical experiment, a ZED2 stereo camera is used to detect the human body, track human skeleton and locate human arms. The visualization of the human skeleton tracking in shown in Fig. \ref{fig:zed}.An auxiliary RGB camera is used to locate blocks with ArUco markers.  

\subsection{Training Result}
\begin{figure}[h]
    \centering
    \subfigure[]{\includegraphics[width=0.235\textwidth]{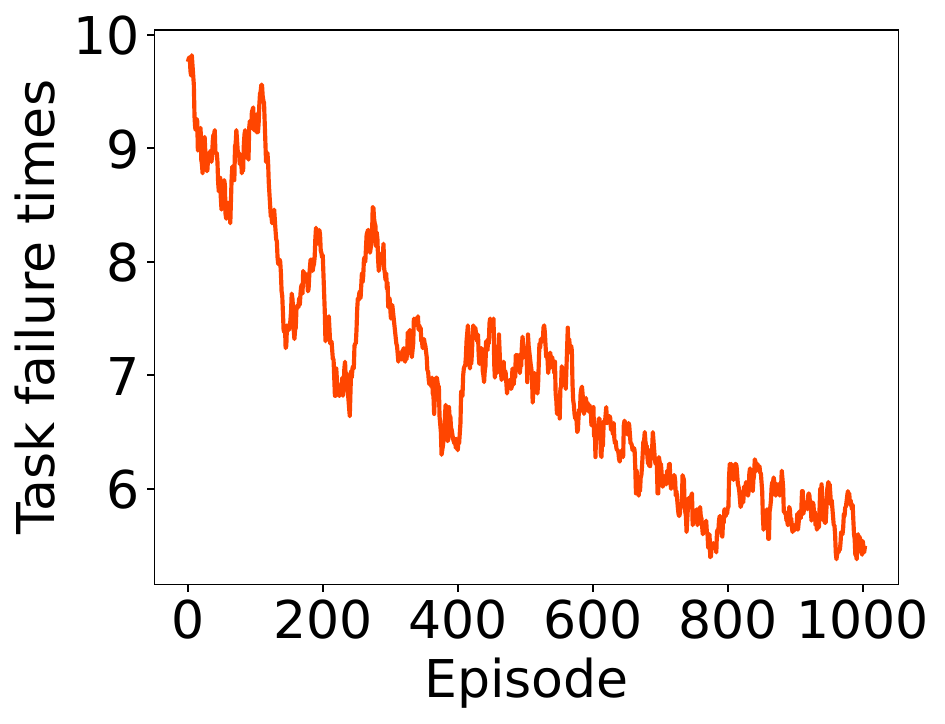}\label{fig:failure_times}} 
    \subfigure[]{\includegraphics[width=0.23\textwidth]{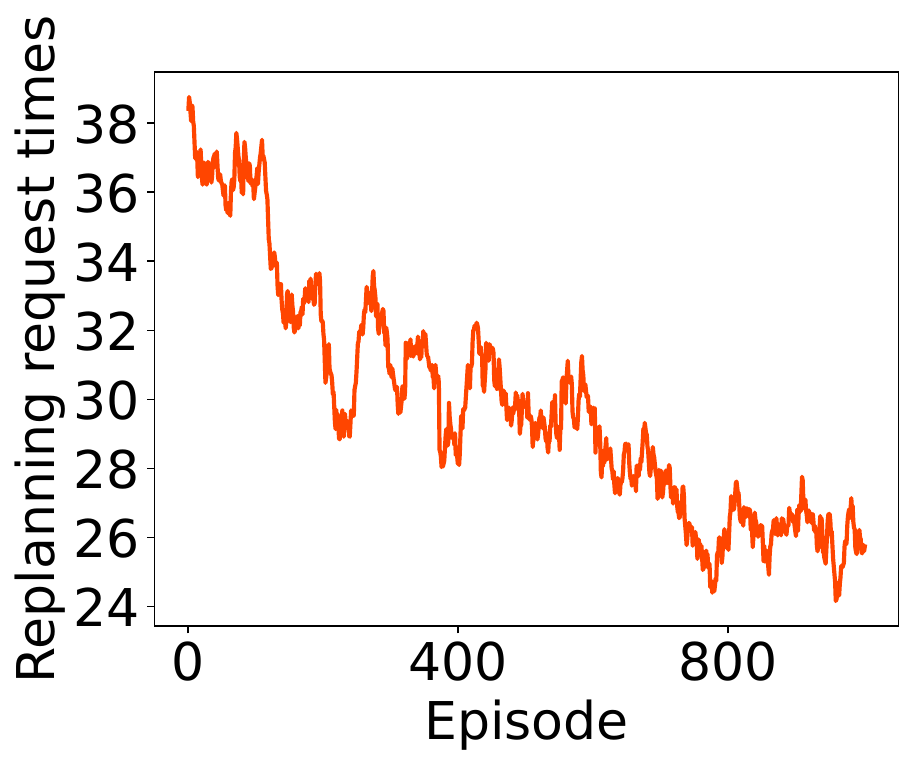}\label{fig:replan_times}} 
    \caption{The RL task planner learning curves. After over 800 episode training, the learning curves converge, both the task failure count and replanning requests are reduced by optimizing accumulated rewards. }
    \label{fig:learning_curves}
\end{figure}

\begin{figure*}[h]
    \centering
    \vspace{2 mm}
    \includegraphics[width=1\textwidth]{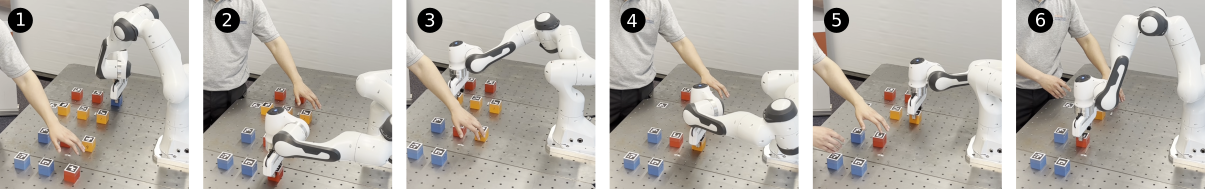}
    \caption{The task sequence with RL task planner. The robot has to pick 6 out of 12 task objects, while the human keep changing the workspace in one HRC process. With RL task planner, the robot adjusts the task sequence (from 1 to 6) to avoid intervention between the human and the robot. The replanning requests and task failure count will also be reduced by applying more rational task sequence.}
    \label{fig:task_real_exp}
\end{figure*}

To illustrate the learning process, we first give the hyperparameters in Table \ref{table: table_ppo_hpp}, and illustrate the learning curves in Fig. \ref{fig:learning_curves}. We use two indices, task failure count and the number of replanning request, to illustrate the improvement of the robot's behaviours. The task failure count indicates how many times the robot chooses the task which are occupied by human (unaccessible) during each episode. The number of replanning requests indicates how many times the robot has to replan its trajectory deal to the human arm intervention.   
The learning curves show that by accumulating the reward, the RL agent can reduce the task failure count and the number of replanning requests. 
After training in the simulation, we implement the trained planner in a real-world scenario. As shown in Fig. \ref{fig:task_real_exp}, the robot chooses tasks which are outside the human congested area. More experiment videos can be found here: \url{https://youtu.be/hgiLHCZw_-M}.

\begin{table}
\vspace{2 mm}
\caption{PPO Hyperparameters}\label{table: table_ppo_hpp}
\centering
\begin{tabular}{ |c|c| }
 \hline
 Name & Value\\
  \hline
 \verb|policy| & MlpPolicy \\
 \hline
 \verb|learning_rate| & 0.0005  \\
 \hline
 \verb|n_steps| & 200  \\
 \hline
 \verb|batch_size| & 50 \\
 \hline
\end{tabular}
\end{table}

\subsection{Comparative Experiment}
\begin{figure}
    \centering
    \subfigure[]{\includegraphics[width=0.235\textwidth]{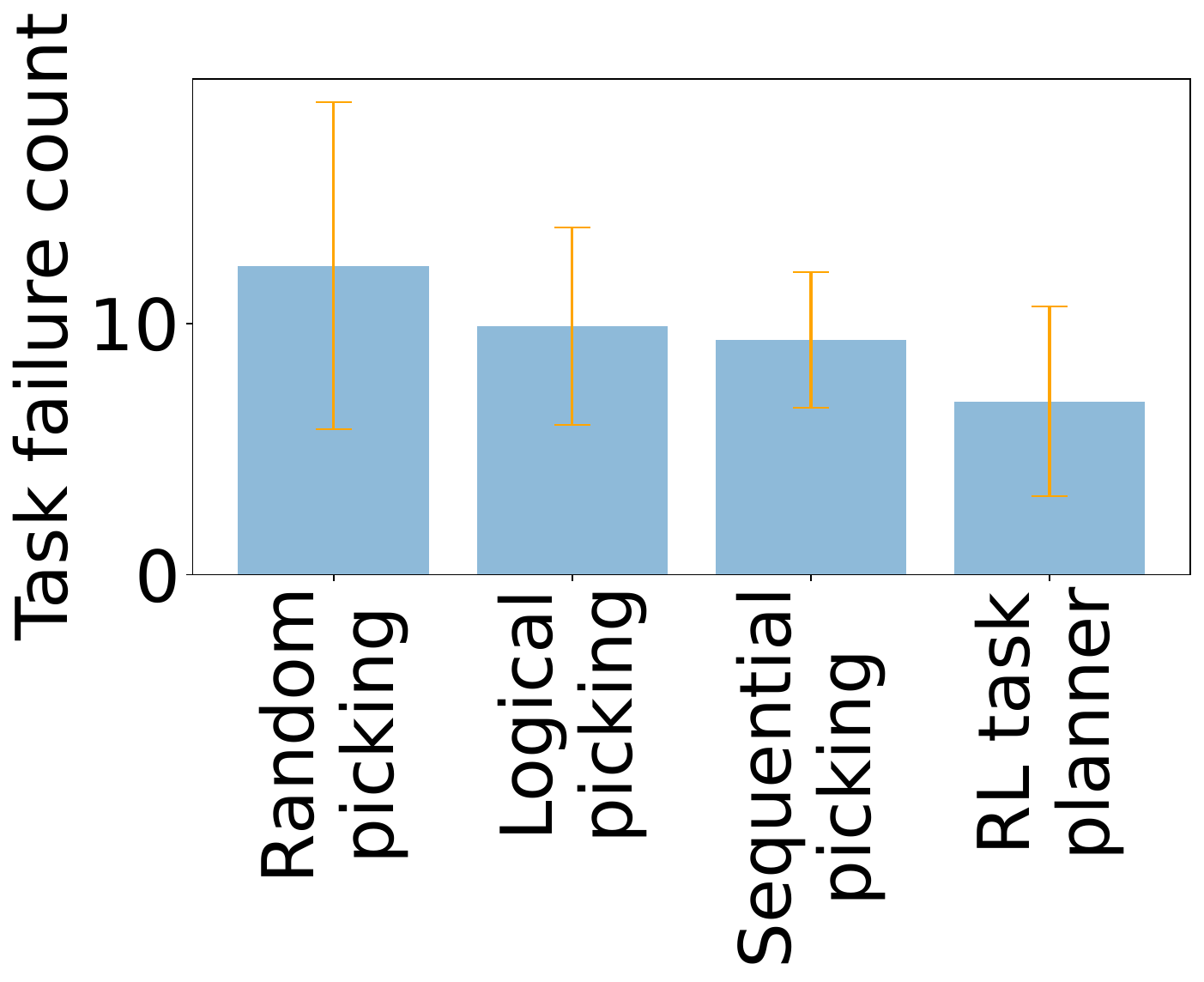}\label{fig:replanner_dilemma_a}} 
    \subfigure[]{\includegraphics[width=0.235\textwidth]{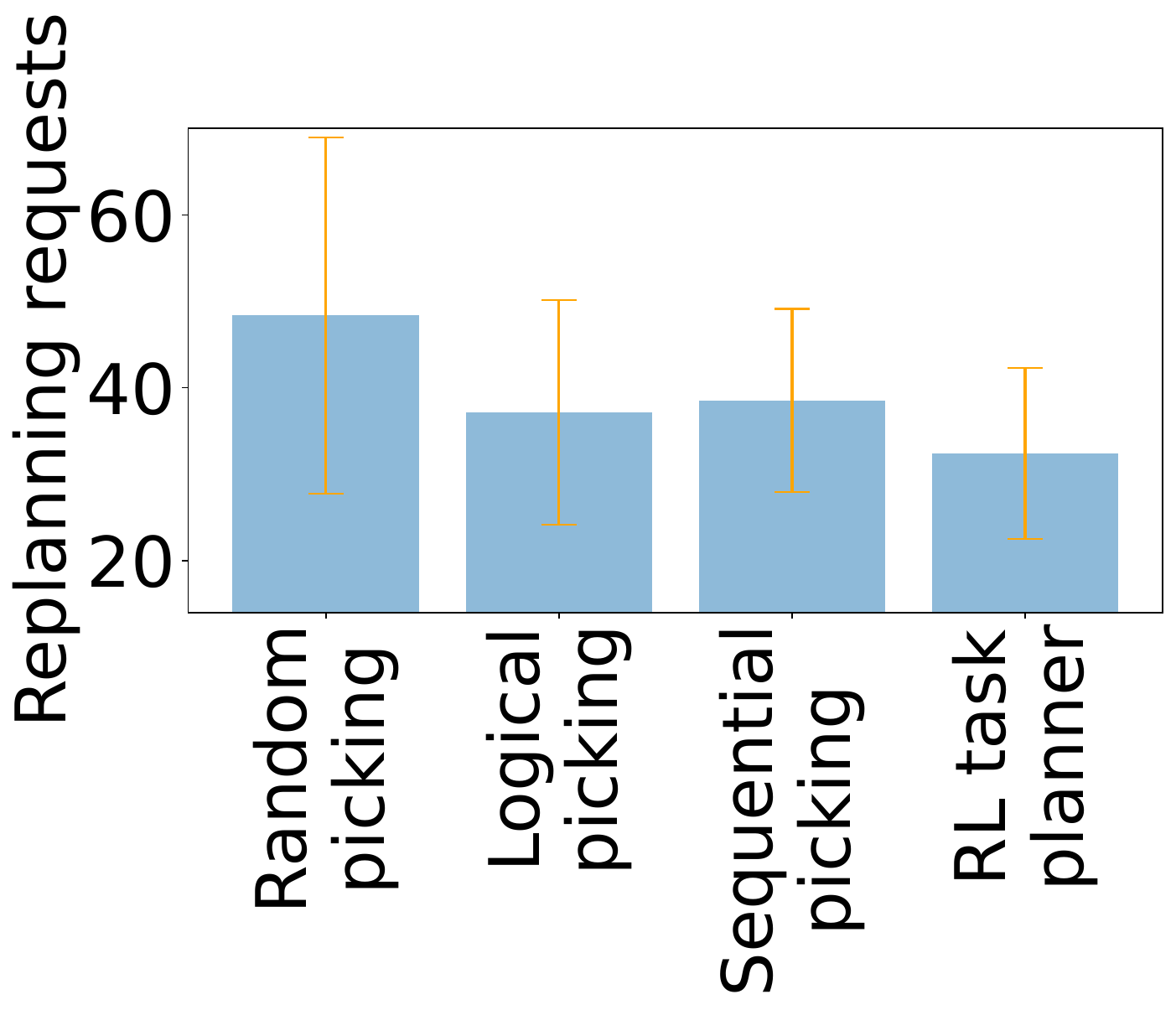}\label{fig:replanner_dilemma_b}} 
    \caption{System performance comparison statistics in the human arm normally distributed scenario. We use different task planning strategies in the same scenario and run the task execution process 20 times with each strategy. The blue bars are the mean values, and the orange lines are the standard deviations of the task failure count and the number of replanning requests, in which the lower value intuitively means more efficient performance.}
    \label{fig:task_compare_n}
\end{figure}

To further validate the method, we compare the proposed RL task planner with three other types of task sequencing logic, i.e. random picking, logical picking and sequential picking, in the same setting explained in \ref{subsec:task_setting}. The random picking represents the system without any task planner, where the agent tries to finish the tasks with no rational sequence. The logical picking chooses the next task following the simple logic (i.e. choose the task which is currently available). The sequential picking represents the hard-coded task planner, where the agent tries to finish the tasks following a fixed sequence. 
During validation, the re-RRT* motion planner will be running to avoid the collisions and count the number of replanning requests. With each task picking strategy, we run the task finishing process 20 times and collect the data of the task failure count and the replanning requests. Intuitively, the less task failure count and replanning requests mean the more efficient performance of the system.  
As the result shown in Fig. \ref{fig:task_compare_n}: (1) the task failure count and the number of replanning requests are related to the task picking strategy, a more rational task picking strategy can reduce the task failure count and the number of replanning requests; (2) The proposed RL task planner has better performance of reducing the task failure count and the replanning requests, thus the efficiency of the HRC system is improved with the RL task-motion planner. 

\begin{figure}
    \centering
    \subfigure[]{\includegraphics[width=0.235\textwidth]{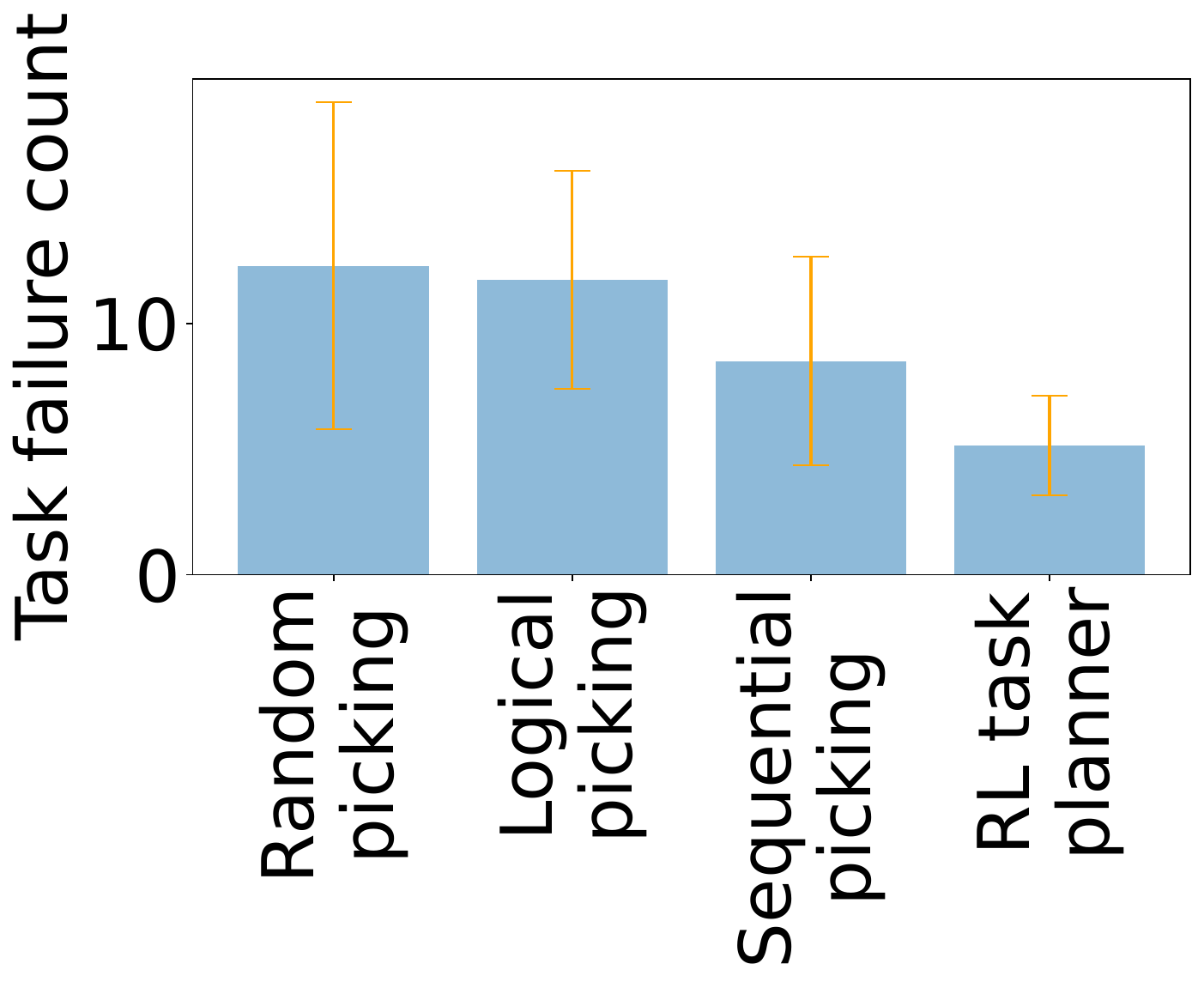}\label{fig:replanner_dilemma_a}} 
    \subfigure[]{\includegraphics[width=0.235\textwidth]{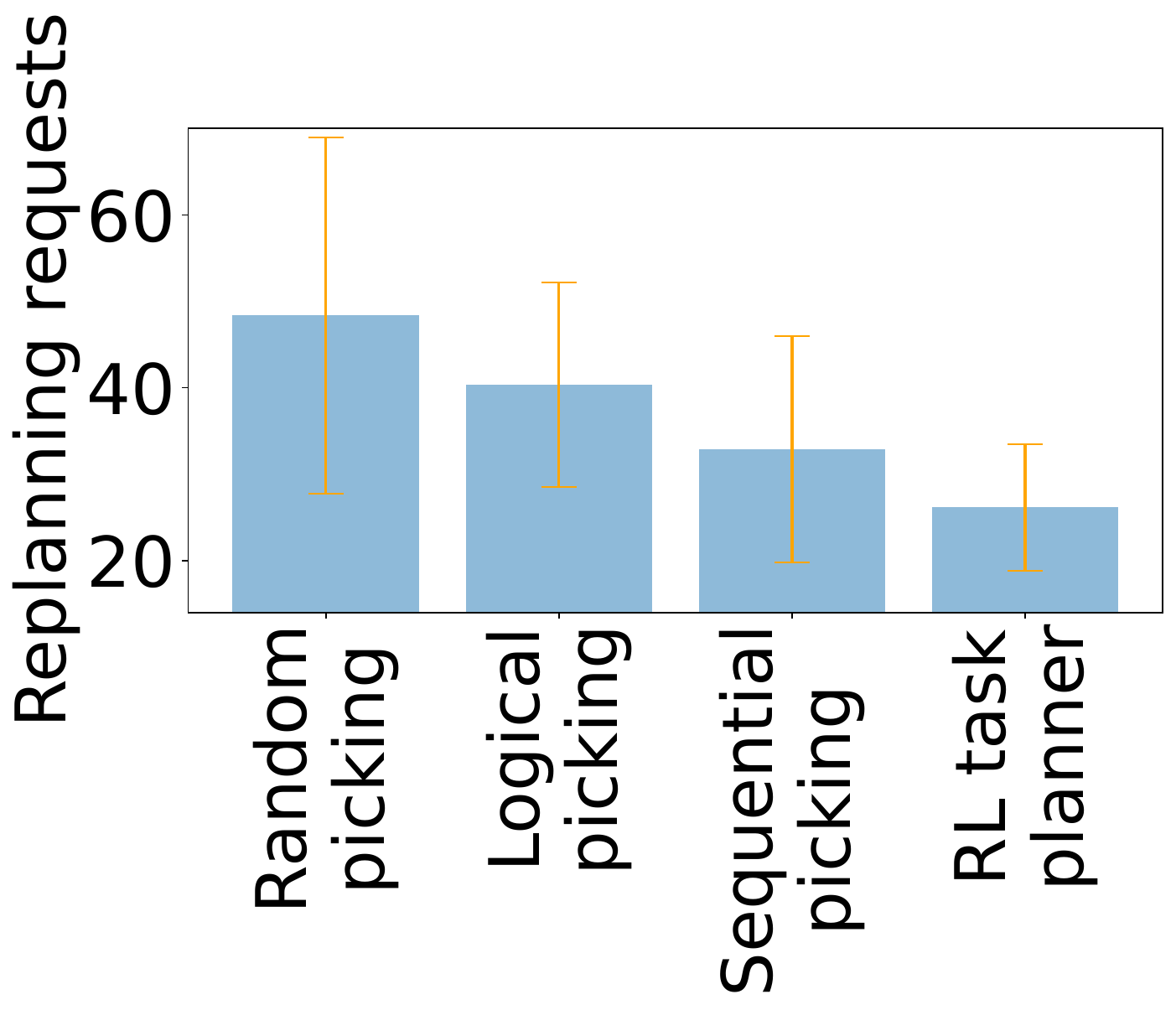}\label{fig:replanner_dilemma_b}} 
    \caption{System performance comparison in the human arm randomly moving scenario. The randomness is increased by making the human arms randomly moving (uniformly distributed) throughout the whole workspace. The blue bars are the mean values, and the orange lines are the standard deviations of the task failure count and the number of replanning requests. The results show that the RL task planner is more robust to the environment changes than the other three task picking strategies.}
    \label{fig:task_compare_u}
\end{figure}

Moreover, to illustrate the robustness of the RL task planner, we change the motion distribution of the human arms from the Gaussian distribution to the uniform distribution, where the human arms randomly move in the whole workspace. The comparison is shown in Fig. \ref{fig:task_compare_u}. The results show that while the sequential picking's performance deteriorates when the uncertainty increase, the performance of the RL task planner is robust to the changes of human arm moving pattern. 

\section{CONCLUSIONS}\label{sec:conclusion}
In this paper, we present a hybrid task-motion planner with a re-RRT* motion planning and an RL task planning. The re-RRT* motion planner avoids human arms by checking the collisions and requesting new paths when the current path is not valid anymore. The RL task planner learns to improve safety and efficiency by choosing more rational task sequences regarding human arm motions. We illustrate that task failure count and the number of replanning requests can be reduced by using the smart task planner. We implement the proposed method on a Franka Emika Panda manipulator in a task finishing scenario and deploy the RL training environment in ROS and Gazebo. The results show that the hybrid RL task-motion planner can improve the task efficiency and the safety. However, the current work only consider the homogeneous tasks and no logical task sequence is considered, therefore, for the future work, the RL planner can be combined with logic programming to solve the more complex situations.

\medskip
\newpage
\bibliographystyle{ieeetr}
\bibliography{reference}

\end{document}